\documentclass[ruled,linesnumbered,journal, a4paper]{IEEEtran}
\usepackage[left=37pt, right=37pt, top=54pt,bottom=114pt]{geometry}

\usepackage[T1]{fontenc}
\usepackage{array}
\usepackage{booktabs}
\usepackage{algorithm2e}
\usepackage{amsmath}
\usepackage{amssymb}
\usepackage{graphicx}
\usepackage{cite}
\usepackage{multirow}
\usepackage[unicode=true,
bookmarks=true,bookmarksnumbered=true,bookmarksopen=true,bookmarksopenlevel=1,
breaklinks=false,pdfborder={0 0 0},pdfborderstyle={},backref=false,colorlinks=false]
{hyperref}
\hypersetup{pdftitle={Your Title},
	pdfauthor={Your Name},
	pdfpagelayout=OneColumn, pdfnewwindow=true, pdfstartview=XYZ, plainpages=false}

\providecommand{\tabularnewline}{\\}

\LinesNumbered
\usepackage[table]{xcolor}
\usepackage{tikz}
\usetikzlibrary{shapes,arrows}
\usepackage[english]{babel}

\tikzstyle{line}=[draw] 
\usepackage{algorithmic}

\usepackage{graphicx}

\makeatother

\begin{document}
	
	\title{Deep Recurrent Electricity Theft Detection in AMI Networks
		with Random Tuning of Hyper-parameters}
	
	\author{\IEEEauthorblockN{Mahmoud Nabil\IEEEauthorrefmark{1},}\ \IEEEauthorblockN{Muhammad~Ismail\IEEEauthorrefmark{2},}\ \IEEEauthorblockN{Mohamed~Mahmoud\IEEEauthorrefmark{1},}\ \IEEEauthorblockN{Mostafa~Shahin\IEEEauthorrefmark{2},}\ \IEEEauthorblockN{Khalid~Qaraqe\IEEEauthorrefmark{2},}\ and \IEEEauthorblockN{Erchin~Serpedin\IEEEauthorrefmark{2}}\ \\\IEEEauthorblockA{\IEEEauthorrefmark{1}Department of Electrical and Computer Engineering,
			Tennessee Tech. University, TN,
			USA}\\\IEEEauthorblockA{\IEEEauthorrefmark{2} Department of
			Electrical and Computer Engineering, Texas A\&M University at Qatar,
			Doha, Qatar}} \maketitle

	\begin{abstract}
		Modern smart grids rely on advanced metering infrastructure (AMI)
		networks for monitoring and billing purposes. However, such an
		approach suffers from electricity theft cyberattacks. Different
		from the existing research that utilizes shallow, static, and
		customer-specific-based electricity theft detectors, this paper
		proposes a generalized deep recurrent neural network (RNN)-based
		electricity theft detector that can effectively thwart these
		cyberattacks. The proposed model exploits the time series nature of
		the customers\textquoteright{} electricity consumption to implement
		a gated recurrent unit (GRU)-RNN, hence, improving the detection
		performance. In addition, the proposed RNN-based detector adopts a
		random search analysis in its learning stage to appropriately fine-tune its hyper-parameters. Extensive test studies are carried out to
		investigate the detector\textquoteright s performance using publicly
		available real data of 107,200 energy consumption days from 200
		customers. Simulation results demonstrate the superior performance
		of the proposed detector compared with state-of-the-art electricity
		theft detectors.
	\end{abstract}
	
	\vspace{-12bp}
	
	\section{Introduction \label{sec:Introduction}}
	
	Electricity theft results in high financial losses for several
	countries such as the United States ($\$6$ billion/year) and India
	($\$17$ billion/year) \cite{jokar2016electricity},
	\cite{singh2017entropy}. Other developing countries lose almost
	$50\%$ of their electricity revenue due to theft
	\cite{antmann2009reducing}. Recently, advanced metering
	infrastructure (AMI) networks are utilized within smart grids for
	monitoring, asset management, and billing purposes. The AMI networks
	rely on smart meters located in the customers\textquoteright{}
	premises to regularly report their energy consumption. This approach has the potential to hinder traditional physical electricity theft attacks
	including line hooking or meter tampering \cite{dpreviewstaff}.
	Nevertheless, a new category of electricity theft attacks that
	target the AMI networks has appeared, namely electricity theft
	cyberattacks. These attacks jeopardize the integrity of the
	customer's energy consumption data as malicious customers hack into
	their smart meters to manipulate their own energy consumption
	values. Despite the difficulties associated with detecting such
	cyberattacks, the customers' fine-grained energy consumption data
	is considered to be a promising tool that can be used to boost
	automated electricity theft detection mechanisms.
	
	Several automated theft detection techniques have been proposed in
	literature
	\cite{nizar2008power,nagi2011improving,ramos2011new,angelos2011detection,lin2014non,amin2015game,zhou2015dynamic,liu2015cyberthreat,jindal2016decision,bhat2016identifying,zhan2016non,tariq2016electricity,xiao2017electricity}.
	These detection techniques can be categorized into three main
	groups: machine learning, state estimation, and game theory-based
	techniques. Compared with these techniques, machine learning-based
	techniques present the most mature and effective mechanism to tackle
	this problem due to their superior detection rate on real datasets \cite{jokar2016electricity}. However, the existing techniques suffer from several limitations. First, the existing detectors employ shallow machine
	learning architectures such as support vector machines (SVMs), while
	deep learning architectures can better capture the behavior of the
	input data resulting in a higher detection rate. In
	addition, the existing detectors rely on static input data \cite{jokar2016electricity,jindal2016decision}, however,
	exploiting the temporal correlation within the time series energy
	consumption data of the customers can further enhance the detection
	performance. Moreover, some of the existing works \cite{jokar2016electricity} adopt customer-specific models, where a detector's model is developed for
	each customer using only its energy consumption data in order to
	detect any future thefts. However, customer-specific models are not
	practical due to several reasons. On one hand, it would be
	impossible to train new detector models for new customers joining
	the system as these new customers do not have any historical
	energy consumption records. Furthermore, customer-specific detectors
	are not robust against contamination attacks, where the customer
	data that is used to train the detector is already malicious yet
	falsely labeled as honest data to confuse the detector.
	
	In this work, our objective is to develop a generalized deep
	recurrent neural network (RNN)-based electricity theft detector. In
	specific, RNNs offer a considerable leverage while working with time
	series data since it is represented as a feedback network where the
	output at the current state is the input for the next state. To the
	best of our knowledge, only \cite{bhat2016identifying} investigates
	the application of deep neural networks for electricity theft
	detection. While a long-short-term-memory (LSTM) architecture of RNNs is investigated in \cite{bhat2016identifying} to tackle the electricity theft detection problem, a comprehensive study is needed to assess the performance of RNN-based detectors on
	real energy consumption data sets. In addition, further
	investigation should be considered in tuning the network
	hyper-parameters to further improve the training time and the
	model's performance.
	
	The contributions of this paper are summarized as follows:
	\begin{itemize}
		\item We propose a generalized RNN-based electricity theft detector using hidden
		gated recurrent unit (GRU) layers. The generality of the model stems from its training on
		the energy consumption load profiles of different residential
		customers' patterns.
		\item We carry out hyper-parameter tuning based on a random search to further
		improve the performance. Random search is proven to outperform grid
		search by finding a good hyper-parameter set in faster execution time
		\cite{bergstra2012random}.
		\item We evaluate the performance of the proposed detector using publicly
		available real data of $107,200$ energy consumption days from $200$
		customers against a set of different types of cyber attacks.
		Simulation results demonstrate the robustness of the proposed model
		against existing detectors.
	\end{itemize}
	The remainder of this paper is organized as follows. Section
	\ref{sec:Related-Work} reviews the related work. Energy consumption
	data is presented in Section \ref{sec:Energy-Consumption-Data}. The
	proposed detection approach is explained in Section
	\ref{sec:Design-of-Electricity}. Simulation results and discussions
	are presented in Section \ref{sec:Numerical-Results-and}. Finally,
	conclusions are drawn in Section \ref{sec:Conclusion}.
	
	\section{Related Work \label{sec:Related-Work}}
	
	Several studies have addressed the problem of detecting electricity
	theft in the literature. Machine learning studies are the most pertinent
	to the proposed work. In \cite{nizar2008power}, the authors compared
	the performance of two classification algorithms based on single
	feedforward neural network, namely, extreme learning machine (ELM),
	and online sequential extreme learning machine (OSELM), against SVM.
	The proposed approach was able to achieve $70\%$ classification rate with ELM.
	Shallow SVM based detector has also been employed in
	\cite{nagi2010nontechnical} and achieved a detection accuracy up to
	$60\%$ on real data set. The performance of this detector was
	further improved in \cite{nagi2011improving} to reach an accuracy of
	$72\%$ using a fuzzy inference system as a post-processing stage. A
	graph-based approach is proposed in \cite{ramos2011new} that uses
	optimum path forest for electricity theft detection and outperforms
	SVM with an accuracy of $89\%$. 
	To the best of our knowledge, the best
	reported performance on a publicly available data set was achieved
	by \cite{jokar2016electricity}, which employs an SVM classifier,
	with an average detection rate of $94\%$ and a false acceptance rate
	of $11\%$.
	
	Deep learning techniques for electricity theft detection are studied
	only in \cite{bhat2016identifying}, where the author presents a
	comparison between different deep learning architectures such as
	convolutional neural networks (CNN), LSTM-RNNs, and stacked
	auto-encoders. Nevertheless, only two types of
	cyberattacks are presented and synthetic data is used to evaluate the performance of the proposed models, which cannot be compared reliably against models
	trained on real consumption data. Moreover, the effect of fine-tuning
	the models hyper-parameters is not studied in
	\cite{bhat2016identifying}.
	
	\begin{table}[t]
		\caption{Cyber Attacks in \cite{jokar2016electricity}. \label{tab:cyberattacks-in}}
		\center%
		\begin{tabular}{>{\centering}p{6cm}}
			\toprule
			Cyber Attack\tabularnewline
			\midrule
			\midrule
			$f_{1}(E_{c}(d,t))=\alpha E_{c}(d,t)$\tabularnewline
			\midrule
			$f_{2}(E_{c}(d,t))=\beta(d,t)E_{c}(d,t)$\tabularnewline
			\midrule
			\scalebox{0.78}{
				
				$f_{3}(E_{c}(d,t))=\begin{cases}
				0 & \forall t\in[t_{\text{\scriptsize{i}}}(d),t_{\text{\scriptsize{f}}}(d)]\\
				E_{c}(d,t) & \forall
				t\notin[t_{\text{\scriptsize{i}}}(d),t_{\text{\scriptsize{f}}}(d)]
				\end{cases}$
				
			}\tabularnewline
			\midrule
			$f_{4}(E_{c}(d,t))=\mathbb{{E}}[E_{c}(d)]$\tabularnewline
			\midrule
			$f_{5}(E_{c}(d,t))=\beta(d,t)\mathbb{{E}}[E_{c}(d)]$\tabularnewline
			\midrule
			$f_{6}(E_{c}(d,t))=E_{c}(d,T-t+1)$\tabularnewline
			\bottomrule
		\end{tabular}

	\end{table}
	\vspace{-3bp}
	\section{Energy Consumption Data \label{sec:Energy-Consumption-Data}}
	
	Define a set of customers $\mathbb{\mathcal{C}}$, a set of days
	$\mathbb{\mathcal{{D}}}=\{1,\dots,D\}$, and a set of periods
	$\mathbb{\mathcal{{T}}}=\{1,\dots,T\}$ of equal duration within each
	day. A smart meter is installed for each customer to regularly
	report the energy consumption for load monitoring and billing. For
	each customer $c$, the energy consumption value at a specific day
	$d$ and time $t$ is denoted as $E_{c}(d,t)$. Hence, each customer
	has an energy consumption record matrix $E_{c}$ of actual
	consumptions, where the rows of $E_{c}$ span the days in
	$\mathbb{\mathcal{{D}}}$, while the columns span the periods in
	$\mathcal{{T}}$. The smart meter reports the consumption to the
	utility for customer $c$ on day $d$ and period $t$ as $R_{c}(d,t)$.
	An honest customer reports the actual energy consumed at the end of
	the consumption period, and hence, $R_{c}(d,t)=E_{c}(d,t)$. On
	contrary, a malicious customer aims to manipulate $E_{c}(d,t)$ to
	lower its electricity bill by modifying the smart meter readings. Hence, the cyberattack is formally defined as
	$R_{c}(d,t)=f(E_{c}(d,t))$, where $f(\cdot)$ denotes a cyberattack
	function that results in a reduced version for the malicious
	customer electricity bill. Although it is easy to collect honest
	samples for a customer $c$ by monitoring its reports to the utility,
	malicious reports might not be easy to collect. This is because, the
	customer reports may all be honest or it has not been detected
	previously as a malicious customer. To tackle this problem, a set of
	cyberattack functions were defined in \cite{jokar2016electricity}
	as indicated in Table \ref{tab:cyberattacks-in}. Each function
	$f(\cdot)$ creates a different attack scenario that aims to reduce
	the customers\textquoteright{} energy consumption $E_{c}(d,t)$. The
	first attack $f_{1}(\cdot)$ aims to reduce $E_{c}(d,t)$ by some
	fraction, where $\alpha$ denotes a flat reduction percentage. On the
	other hand, attack $f_{2}(\cdot)$ dynamically reduces $E_{c}(d,t)$
	by a value controlled by the time function $\beta(d,t)$. The third
	attack $f_{3}(\cdot)$ represents a selective time filtering
	function, where the malicious customer reports zero reading during
	the interval
	$[t_{\text{\scriptsize{i}}}(d),t_{\text{\scriptsize{f}}}(d)]$,
	otherwise, the customer reports the actual consumption $E_{c}(d,t)$.
	Here, $t_{\text{\scriptsize{i}}}(d)$ and
	$t_{\text{\scriptsize{f}}}(d)$ denotes the initial and the final
	periods of the interval. The next two attacks $f_{4}(\cdot)$ and
	$f_{5}(\cdot)$ are based on the expected value of the energy
	consumption for the malicious customer for a given day denoted by
	$\mathbb{{E}}[E_{c}(d)]$. In $f_{4}(\cdot)$ , the attacker reports a
	flat value during the day, while in $f_{5}(\cdot)$ the attacker
	reduces $\mathbb{{E}}[E_{c}(d)]$ dynamically from time to time using
	function $\beta(d,t)$. The last attack $f_{6}(\cdot)$ is a reverse
	function that reorders the energy consumption reports during the day
	so that higher reports are assigned to low tariff periods.
	
	\begin{figure}[t]
		\center
		
		\includegraphics[scale=0.45]{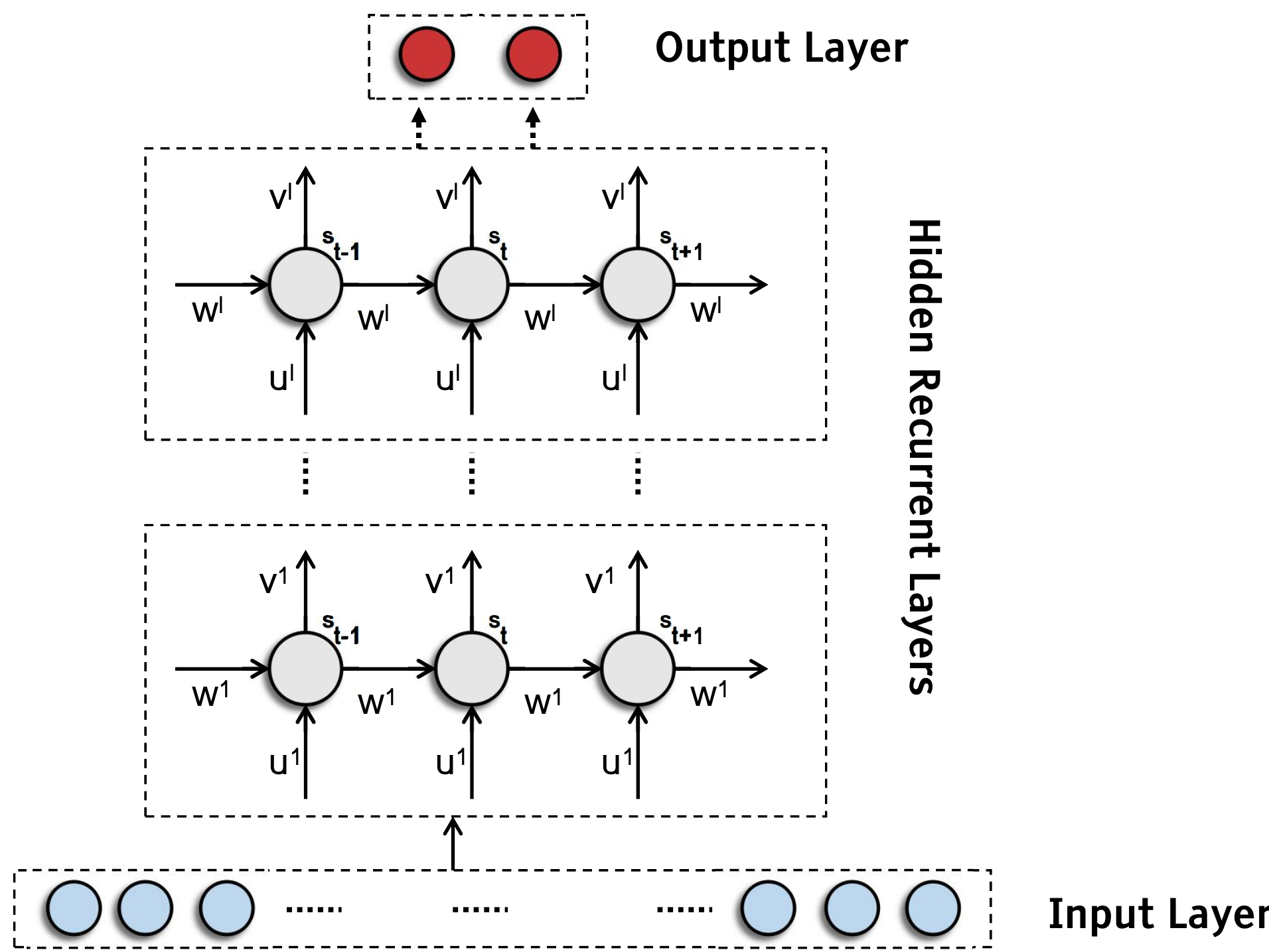}
		
		\caption{Architecture of general RNN-based electricity theft detector. \label{fig:Architecture-of-general}}
	\end{figure}
	
	Applying the aforementioned cyber attacks on the energy consumption
	record matrix $E_{c}$ results in six matrices $M_{c,i}$, where
	$i\in\{1,\dots,6\}$, for each attack $f_{i}(\cdot)$. The complete
	data set for a customer is denoted as $\hat{X}_{c}$, which is a
	concatenation for $E_{c}$ and all malicious matrices $M_{c,i}$. In
	addition, each row $X_{c}(d,\cdot)$ represents either an honest day
	sample (i.e, $E_{c}$ samples) labeled with 0 or a malicious sample
	(i.e., $M_{c,i}$ samples) labeled with 1. Since $\hat{X}_{c}$ is an
	imbalanced dataset where the ratio of the honest samples to the
	malicious samples is 1:6, the detector may be biased towards the
	dominant class category (i.e., malicious samples). To mitigate the
	bias effect, the adaptive synthetic sampling approach (ADASYN)
	\cite{he2008adasyn} is used to oversample the minor class so that
	the ratio between the honest and the malicious samples is almost the
	same. Each data set $\hat{X}_{c}$ is partitioned into two disjoints
	data sets for model evaluation, specifically, a training set
	$\hat{X}_{c,\text{\sc{tr}}}$ and atesting set
	$\hat{X}_{c,\text{\sc{tst}}}$ with ratio 3:2. The training sets
	of all customers in $\mathbb{\mathcal{C}}$ are merged together to
	form $\hat{X}_{\text{\sc{tr}}}$. Similarly, the test sets of all
	customers are merged together to form $\hat{X}_{\text{\sc{tst}}}$.
	Hence, all customers have the same opportunity in training and
	evaluating the general model. As different customers have different
	ranges of energy consumption values during different report periods
	$\mathbb{\mathcal{{T}}}$, different consumption periods have
	different opportunities to influence the optimization algorithm for
	the final model. Hence, in order to enforce equal opportunities for
	different consumption periods, feature scaling is applied to
	$\hat{X}_{\text{\sc{tr}}}$. The mean and the variance of
	$\hat{X}_{\text{\sc{tr}}}$ are calculated and used to scale
	$\hat{X}_{\text{\sc{tr}}}$ so that each consumption period would
	have zero mean and unit standard deviation. The output of the
	feature scaling stage is denoted as $X_{\text{\sc{tr}}}$. The same
	scaling scores are further applied to the test set to get
	$X_{\text{\sc{tst}}}$.
	
	To evaluate the performance of the detector, three metrics are used
	as follows. The detection rate (DR) measures the percentage of the
	correctly detected malicious attacks. The false acceptance (FA) rate
	measures the percentage of the honest samples that are falsely
	identified as malicious. The highest difference (HD) measures the
	difference between DR and FA \cite{jokar2016electricity} . The
	detection latency is ignored as electricity theft attacks do not
	result in an immediate loss to the utility. Moreover, malicious
	customers can be fined later for any detected thefts.
	
	\begin{algorithm}[t]
		\SetAlgoLined \KwData{$X_{\text{\sc{tr}}}$} \KwResult{Optimal
			parameters $U_{(.)}^{l}$,$W_{(.)}^{l}$,$V_{(.)}^{l}$, and
			$b_{(.)}^{l}$ $\forall l\in[1, \hdots, L]$}
		
		\textbf{Initialization: }Weights
		$U_{(.)}^{l}$,$W_{(.)}^{l}$,$V_{(.)}^{l}$, and $b_{(.)}^{l}$
		$\forall l\in[1, \hdots, L]$, $i=1$
		
		\While{$i\neq I$} { Initialize: $m=1$\\
			\While{$m\neq M$} {
				
				\For{each training example $x$ in mini-batch $m$} {
					
					\textbf{Feed Forward:}
					
					\For{each recurrent layer $l$} { \For{each time step $t$}{
							
							$z_{t}^{l}=\sigma(o_{t}^{l-1}U_{z}^{l}+s_{t-1}^{l}W_{z}^{l}+b_{z}^{l})$
							
							$r_{t}^{l}=\sigma(o_{t}^{l-1}U_{r}^{l}+s_{t-1}^{l}W_{r}^{l}+b_{r}^{l})$
							
							$h_{t}^{l}=\tanh(o_{t}^{l-1}U_{h}^{l}+(s_{t-1}^{l}\odot
							r^{l})W_{h}^{l}+b_{h}^{l})$
							
							$s_{t}^{l}=(1-z^{l})\odot h^{l}+z^{l}\odot s_{t-1}^{l}$
							
							$o_{t}^{l}=\text{softmax}(V^{l}s_{t}^{l}+b_{o}^{l})$
							
						}
						
					}
					
					\textbf{Back propagation: }Compute $\nabla_{U_{(.)}^{l}}C(x)$, $\nabla_{V_{(.)}^{l}}C(x)$,
					$\nabla_{W_{(.)}^{l}}C(x)$, and $\nabla_{b_{(.)}^{l}}C(x)$
					
				}
				
				\textbf{Weight and bias update: }
				
				$U_{(.)}^{l}=U_{(.)}^{l}-\frac{\eta}{K}\sum_{x}\nabla_{U_{(.)}^{l}}C(x)$
				
				$V_{(.)}^{l}=V_{(.)}^{l}-\frac{\eta}{K}\sum_{x}\nabla_{V_{(.)}^{l}}C(x)$
				
				$W_{(.)}^{l}=W_{(.)}^{l}-\frac{\eta}{K}\sum_{x}\nabla_{W_{(.)}^{l}}C(x)$
				
				$b_{(.)}^{l}=b_{(.)}^{l}-\frac{\eta}{K}\sum_{x}\nabla_{b_{(.)}^{l}}C(x)$
				
			}
			
		}
		
		\caption{Electricity Theft Detector Training \label{alg:Electricity-Theft-Detector}}
	\end{algorithm}
	\vspace{-8bp}
	\section{Design of Electricity Theft Detector \label{sec:Design-of-Electricity}}
	
	This section describes the two stages used to develop the RNN-based
	general detector. The first is a training stage that describes the
	detector architecture and parameters, and the second is the
	hyper-parameter optimization stage that uses a random search-based
	approach for fine-tuning.
	
	\subsection{Training Stage}
	
	\begin{algorithm}[t]
		\SetAlgoLined \KwData{$X_{\text{\sc{tr}}}$} \KwResult{Optimal hyper
			parameters $L^{*}$,$N^{*}$,$O^{*}$, $A_{H}^{*}$, and $A_{O}^{*}$}
		
		\textbf{Initialization: }Weights
		$U_{(.)}^{l}$,$W_{(.)}^{l}$,$V_{(.)}^{l}$, and $b_{(.)}^{l}$
		$\forall l\in[1, \hdots, L]$, $i=1$
		
		\While{$i\neq I$} {
			
			$L[i]\leftarrow\mathcal{L}$ , $N[i]\leftarrow\mathcal{N}$ ,$O[i]\leftarrow\mathcal{O}$,
			$A_{H}[i]\leftarrow\mathcal{A}_{H}$, $A_{O}[i]\leftarrow\mathcal{A}_{O}$
			
			\For{ each $\bar{X}_{\text{\sc{tr}}}$ , $\bar{X}_{\text{\sc{tst}}}$
				in k-folds$(X_{\text{\sc{tr}}})$} {
				
				Apply Algorithm 1 using sampled hyper-parameters and record DR, FA
				and accuracy.
				
			}
			
			Record average values across folds DR{[}$i${]}, FA{[}$i${]} and
			accuracy{[}$i${]}.
			
		}
		
		Report the hyperparameters of the top three models.
		
		\caption{Random Search Hyper Parameters Optimization \label{alg:Random-Search-Hyper}}
	\end{algorithm}
	
	The architecture of the RNN-based general detector is shown in Fig.
	\ref{fig:Architecture-of-general}. The input layer is denoted by
	vector $x_{c}(d)$, which represents the consumption vector of
	customer $c$ during day $d$ at different reporting periods
	$t\in\mathcal{{T}}$. The RNN architecture consists of $L$ hidden GRU
	layers each with $N$ neurons. Each GRU layer except the last one
	accepts a sequence vector as input and produces a sequence output
	vector. Unlike traditional feedforward layers, recurrent layers are
	more efficient in exploiting patterns from sequential information. The
	output layer has two neurons that define the malicious and the honest
	classes, where the true label can be represented by one-hot vector
	$y(x_{c}(d))=(0\,\,1)^{T}$ for honest customers, and
	$y(x_{c}(d))=(1\,\,0)^{T}$ for malicious customers.
	
	The output vector from each layer $l\in[1,\hdots,L]$ is denoted as
	$o^{l}$, where $o^{1}=x(d)$. Each hidden GRU layer
	$l\in[2,\hdots,L-1]$ has the following parameters:
	\begin{itemize}
		\item The input at time step $t$ is $o_{t}^{l-1}$, which results from the
		previous layer $l-1$.
		\item The hidden state $s_{t}^{l}$ at time step $t$ that represents the
		memory computed based on the previous hidden state $s_{t-1}^{l}$
		of the same layer.
		\item The update gate $z_{t}^{l}=\sigma(o_{t}^{l-1}U_{z}^{l}+s_{t-1}^{l}W_{z}^{l}+b_{z}^{l})$,
		which determines a combination between the new input $o_{t}^{l-1}$
		and the previous memory $s_{t-1}^{l}$. Here, $U_{z}^{l}$ and
		$W_{z}^{l}$ are learnable weights, $\sigma(.)$ is the activation
		function, and $b_{z}^{l}$ is the bias vector.
		\item The reset gate $r_{t}^{l}=\sigma(o_{t}^{l-1}U_{r}^{l}+s_{t-1}^{l}W_{r}^{l}+b_{r}^{l})$,
		which determines the amount of the previous memory $s_{t-1}^{l}$
		that will contribute to the next state using the equation
		$h_{t}^{l}=\tanh(o_{t}^{l-1}U_{h}^{l}+(s_{t-1}^{l}\odot
		r^{l})W_{h}^{l}+b_{h}^{l})$ . Here, $U_{r}^{l}$, $W_{r}^{l}$,
		$U_{h}^{l}$, and $W_{h}^{l}$ are learnable weight matrices,
		$b_{r}^{l}$ and $b_{h}^{l}$ are the bias vectors, and $\odot$ is the
		Hadamard product.
		\item The next state is simply $s_{t+1}^{l}=(1-z^{l})\odot h^{l}+z^{l}\odot s_{t}^{l}$
		, and the output at time $t+1$ is
		$o_{t+1}^{l}=\text{softmax}(V^{l}s_{t+1}^{l}+b_{o}^{l})$, where
		$V^{l}$ is a learnable weight matrix.
	\end{itemize}
	The detector training stage involves learning the parameters
	$U_{(.)}^{l}$,$W_{(.)}^{l}$,$V_{(.)}^{l}$, and $b_{(.)}^{l}$ for all
	layers, which results in the desired output vector $y(x_{c}(d))$ for
	any input $x_{c}(d)$. The optimization objective to find these
	weights can be achieved by minimizing the cross-entropy cost function
	$C$ defined as
	\begin{equation}
	\begin{aligned}
	\underset{\Theta}{\min}\frac{-1}{S}\sum_{X_{\text{\sc{tr}}}} & \{y^{\text{\sc{t}}}(x_{c}(d))\ln(o^{L}) + \\
	& (1-y^{\text{\sc{t}}}(x_{c}(d)))\ln(1-o^{L})\}, \label{eq:1}
	\end{aligned}
	\end{equation}
	Where $\Theta$ denotes the RNN parameters
	$U_{(.)}^{l}$,$W_{(.)}^{l}$,$V_{(.)}^{l}$, and $b_{(.)}^{l}$ for all
	layers and $S$ represents the total number of the training samples
	for all customers, which equals to the number of rows in
	$X_{\text{\sc{tr}}}$. Solving \eqref{eq:1} for the optimal
	parameters $\Theta$ is done using an iterative gradient
	descent-based optimization algorithm. This is achieved by
	partitioning $X_{\text{\sc{tr}}}$ into $M$ mini-batches of
	equal size and executing Algorithm
	\ref{alg:Electricity-Theft-Detector} for $I$ total iterations. Each
	iteration $i$ has two main stages, namely, feed forward and
	back-propagation. In the feed forward stage, the training samples in
	the mini-batch are passed through all the network layers to
	calculate the predicted output vectors. On the other hand, the back
	propagation stage uses the mini-batches to calculate the gradient of
	the cost function in \eqref{eq:1} with respect to the network
	weights \cite{goodfellow2016deep}. Thereafter, the computed
	gradients are used to modify the weights and biases for each
	iteration. Note that, the back propagation within each layer is
	basically a back propagation through time (BPTT) as each layer is a
	GRU. In addition, $\nabla_{a}$ in Algorithm
	\ref{alg:Electricity-Theft-Detector} denotes the partial derivative
	with respect to $a$. Algorithm \ref{alg:Electricity-Theft-Detector}
	uses stochastic gradient descent (SGD) for optimization with
	learning rate $\eta$.
	
	\subsection{Hyper-parameters Optimization Stage}
	
	\begin{table*}[t]
		\caption{Sampled hyper-parameters results of Algorithm \ref{alg:Random-Search-Hyper}
			using k-fold cross validation\label{tab:The-sampled-results}.}
		\center
		\resizebox{0.7\textwidth}{0.32\textwidth}{
			\begin{tabular}{cccccccccc}
				\toprule 
				$L$  & $N$  & $A_{H}$  & $A_{O}$  & $O$  &  & DR  & FA  & HD  & Accuracy\tabularnewline
				\midrule
				\midrule 
				\multirow{11}{*}{2 } & 201  & Hard Sigmoid  & Sigmoid  & Adam  &  & 0.930  & 0.029 & 0.900  & 0.950\tabularnewline
				\cmidrule{2-10} 
				& 384  & Sigmoid  & Sigmoid  & SGD  &  & 0.707 & 0.313 & 0.394 & 0.697\tabularnewline
				\cmidrule{2-10} 
				& \cellcolor{gray!30}\textbf{428 }  & \cellcolor{gray!30}\textbf{Sigmoid }  & \cellcolor{gray!30}\textbf{Softmax }  & \cellcolor{gray!30}\textbf{Adamax }  & \cellcolor{gray!30} & \cellcolor{gray!30}\textbf{0.946}  & \cellcolor{gray!30}\textbf{0.036}  & \cellcolor{gray!30}\textbf{0.910} & \cellcolor{gray!30}\textbf{0.955}\tabularnewline
				\cmidrule{2-10} 
				& 270  & Relu  & Sigmoid  & Nadam  &  & 0  & 0  & 0  & 0.499 \tabularnewline
				\cmidrule{2-10} 
				& 306  & Hard Sigmoid  & Softmax  & SGD  &  & 0.727 & 0.324 & 0.403 & 0.701 \tabularnewline
				\cmidrule{2-10} 
				& 258  & Hard Sigmoid  & Sigmoid  & SGD  &  & 0.713 & 0.311 & 0.402 & 0.701\tabularnewline
				\cmidrule{2-10} 
				& 154  & Hard Sigmoid  & Softmax  & Adagrad  &  & 0.880 & 0.067  & 0.812 & 0.906\tabularnewline
				\cmidrule{2-10} 
				& 250  & Tanh  & Sigmoid  & Adadelta  &  & 0.916 & 0.051 & 0.865 & 0.932\tabularnewline
				\cmidrule{2-10} 
				& 351  & Sigmoid  & Softmax  & Adamax  &  & 0.941  & 0.035 & 0.905  & 0.952 \tabularnewline
				\cmidrule{2-10} 
				& 284  & Hard Sigmoid  & Softmax  & SGD  &  & 0.739  & 0.340 & 0.398  & 0.699 \tabularnewline
				\cmidrule{2-10} 
				& 105  & Tanh  & Sigmoid  & Adagrad  &  & 0.877 & 0.083 & 0.793 & 0.896\tabularnewline
				\midrule
				\midrule 
				\multirow{9}{*}{3} & 385  & Tanh  & Sigmoid  & Adadelta  &  & 0.928  & 0.037 & 0.890 & 0.945\tabularnewline
				\cmidrule{2-10} 
				& 307  & Sigmoid  & Sigmoid  & RMSprop  &  & 0.923 & 0.031 & 0.891 & 0.945\tabularnewline
				\cmidrule{2-10} 
				& 239  & Tanh  & Sigmoid  & Adagrad  &  & 0.609 & 0.328  & 0.281  & 0.640\tabularnewline
				\cmidrule{2-10} 
				& 272  & Hard Sigmoid  & Sigmoid  & Adadelta  &  & 0.885 & 0.071 & 0.813 & 0.906 \tabularnewline
				\cmidrule{2-10} 
				& 110  & Sigmoid  & Softmax  & Nadam  &  & 0.928 & 0.036 & 0.891 & 0.945 \tabularnewline
				\cmidrule{2-10} 
				& 251  & Tanh  & Sigmoid  & Nadam  &  & 0  & 0  & 0  & 0.499\tabularnewline
				\cmidrule{2-10} 
				& \cellcolor{gray!30}\textbf{310 }  & \cellcolor{gray!30}\textbf{Hard Sigmoid }  & \cellcolor{gray!30}\textbf{Softmax }  & \cellcolor{gray!30}\textbf{Adam }  & \cellcolor{gray!30} & \cellcolor{gray!30}\textbf{0.939}  & \cellcolor{gray!30}\textbf{0.033} & \cellcolor{gray!30}\textbf{0.905}  & \cellcolor{gray!30}\textbf{0.952 }\tabularnewline
				\cmidrule{2-10} 
				& 407  & Hard Sigmoid  & Sigmoid  & Adagrad  &  & 0.867 & 0.075 & 0.791  & 0.895\tabularnewline
				\cmidrule{2-10} 
				& 193  & Sigmoid  & Sigmoid  & Adadelta  &  & 0.875 & 0.068  & 0.807 & 0.903\tabularnewline
				\midrule
				\midrule 
				\multirow{10}{*}{4 } & 492  & Relu  & Sigmoid  & SGD  &  & 0  & 0  & 0  & 0.499\tabularnewline
				\cmidrule{2-10} 
				& 198  & Tanh  & Softmax  & RMSprop  &  & 0.419 & 0.186  & 0.232 & 0.615\tabularnewline
				\cmidrule{2-10} 
				& 352  & Sigmoid  & Softmax  & Adamax  &  & 0.931 & 0.026 & 0.905 & 0.952\tabularnewline
				\cmidrule{2-10} 
				& \cellcolor{gray!30}\textbf{215 }  & \cellcolor{gray!30}\textbf{Tanh }  & \cellcolor{gray!30}\textbf{Softmax }  & \cellcolor{gray!30}\textbf{Adamax }  & \cellcolor{gray!30} & \cellcolor{gray!30}\textbf{0.942}  & \cellcolor{gray!30}\textbf{0.028}  & \cellcolor{gray!30}\textbf{0.914 }  & \cellcolor{gray!30}\textbf{0.957}\tabularnewline
				\cmidrule{2-10} 
				& 275  & Tanh  & Softmax  & Adam  &  & 0  & 0  & 0  & 0.499 \tabularnewline
				\cmidrule{2-10} 
				& 202  & Relu  & Sigmoid  & Adagrad  &  & 0  & 0  & 0  & 0.499\tabularnewline
				\cmidrule{2-10} 
				& 393  & Sigmoid  & Sigmoid  & RMSprop  &  & 0.939 & 0.044 & 0.894 & 0.947\tabularnewline
				\cmidrule{2-10} 
				& 165  & Sigmoid  & Softmax  & Adagrad  &  & 0.898  & 0.073  & 0.825  & 0.912\tabularnewline
				\cmidrule{2-10} 
				& 390  & Hard Sigmoid  & Sigmoid  & Adadelta  &  & 0.608  & 0.359 & 0.249  & 0.624\tabularnewline
				\cmidrule{2-10} 
				& 310  & Tanh  & Sigmoid  & RMSprop  &  & 0.187 & 0.126 & 0.061  & 0.530\tabularnewline
				\bottomrule
			\end{tabular}} 
		\end{table*}
		
		Tuning the hyper-parameters of the detector is a challenging and
		a time-consuming task, however, optimal parameters improve the
		detector performance. Since an exhaustive search on hyper-parameters
		is practically infeasible due to the incurred high computational
		complexity, random search is used to find a sub-optimal but
		an efficient solution in a reduced time compared with the exhaustive
		grid search \cite{bergstra2012random}. In this paper, random search
		is used to tune the following hyper-parameters: the network
		architecture in terms of the number of hidden layers and the number
		of neurons within each hidden layer, the type of activation
		functions used at the RNN layers, the type of activation functions
		used at the output layer, and the type of gradient descent-based
		optimization algorithm used in finding the optimal RNN parameters
		(weights and biases) \cite{goodfellow2016deep}. Denote $\mathcal{L}$
		and $\mathcal{N}$ as two uniform distributions that represent the
		number of hidden layers and neurons to be sampled by the random
		search algorithm. In addition, Let $\mathcal{O}$ represent a uniform
		distribution of optimization algorithms that can include Adam, SGD,
		Adamax, etc. \cite{goodfellow2016deep}. Moreover, two uniform
		distributions are defined to represent the set of activation
		functions used by the hidden and output layers, namely,
		$\mathcal{A}_{H}$ and $\mathcal{A}_{O}$, respectively. Algorithm
		\ref{alg:Random-Search-Hyper} shows the random search over
		$X_{\text{\sc{tr}}}$ for the selection of the optimal
		hyper-parameters defined as $P^{*}$. The maximum number of search iterations is denoted as $I$, where at each iteration all the
		distributions are sampled uniformly and the sampled model is
		evaluated using K-fold cross-validation.
		
		\vspace{-7bp}
		
		\section{Numerical Results and Discussion \label{sec:Numerical-Results-and} }
		
		To evaluate the performance of the proposed RNN-based general
		electricity theft detector, real smart meter data from the Irish
		Smart Energy Trials is used \cite{dataset}. The data set was
		published by the Electric Ireland and Sustainable Energy Authority
		of Ireland in January 2012. It contains the energy consumption
		readings for $5000$ customers over $536$ days between $2009-2010$.
		The customers report their readings every 30 minutes (i.e., the RNN
		input size is $48$), and hence, the total number of reports per
		customer is $25,728$. Daily reports from $200$ customers with
		$107,200$ days are considered as honest energy consumption records,
		which are then used to launch the cyberattacks
		given in Table \ref{tab:cyberattacks-in} for each customer. For cyberattacks
		$f_{1}(\cdot)$, $f_{2}(\cdot)$, and $f_{5}(\cdot)$, $\alpha$ and
		$\beta(d,t)$ are random variables that are uniformly distributed
		over the interval $[0.1,0.8]$ \cite{jokar2016electricity}. For
		attack $f_{3}(\cdot)$, $t_{\text{\scriptsize{i}}}(d)$ is a uniform
		random variable in $[0,42]$, and the duration of the attack, i.e.,
		$t_{\text{\scriptsize{f}}}(d)-t_{\text{\scriptsize{i}}}(d)$, is a
		uniform random variable in $[8,48]$, and hence, the maximum value of
		$t_{f}(d)=48$. Applying all cyberattacks for each customer results
		in $\hat{X}_{c}$ data set which contains $536$ honest samples (days)
		and $3,216$ malicious samples. Each sample in $\hat{X}_{c}$ has $48$
		energy consumption values. For each $\hat{X}_{c}$ data set,
		over-sampling is performed to balance the size of honest and
		malicious classes. Consequently, the total size of each
		$\hat{X}_{c}$ data set contains a total of $6,432$ honest and
		malicious samples. The total number of samples for all $c\in
		\mathcal{C}$ is $1,286,400$. Each $\hat{X}_{c}$ data set is
		partitioned into training set $\hat{X}_{c,\text{\sc{tr}}}$ and
		testing set $\hat{X}_{c,\text{\sc{tst}}}$ with ratio 3:2. The
		training sets for all customers are merged together to form
		$\hat{X}_{\text{\sc{tr}}}$. Similarly, the test sets for all
		customers are merged together to form $\hat{X}_{\text{\sc{tst}}}$.
		Feature scaling on training data produces $X_{\text{\sc{tr}}}$
		and applying the same scores on the test data produces
		$X_{\text{\sc{tst}}}$.
		
		In Algorithm \ref{alg:Electricity-Theft-Detector}, the total number
		of epochs $I=10$ and the batch size is $350$. For Algorithm
		\ref{alg:Random-Search-Hyper}, the options used for the random
		search are the following: $\mathcal{L}=\{2,3,4\}$,
		$\mathcal{N}=[100,500]$, $\mathbb{\mathbb{\mathcal{O}=}}$\{SGD,
		Adadelta, Adagrad , Adam, Adamax, Nadam, RMSprop\},
		$\mathcal{{A}}_{H}=$\{Sigmoid, Relu, Hard Sigmoid, Tanh\}, and
		$\mathcal{{A}}_{O}=$\{Sigmoid, Softmax\}. In addition, the total
		number of iterations is $I=30$, the drop out rate is set to $0.2$
		and the initialization of the network neurons is set to Glorot
		Uniform \cite{goodfellow2016deep}. All the experiments were
		performed on the high-performance cluster (HPC) of the Tennessee Tech University  using two NVIDIA Tesla K80 GPUs. Tensorflow and Python Keras Library are used
		for the implementation.
		
		Table \ref{tab:The-sampled-results} gives the random search
		evaluation results obtained by applying Algorithm
		\ref{alg:Random-Search-Hyper} on the sampled hyper-parameters. The
		results give the average detection performance over K-fold cross-validation (with $K=3$) on $X_{\text{\sc{tr}}}$ in terms of DR, FA,
		and HD. As indicated in the table, the proposed general electricity
		theft detector benefits from the recurrent deep architecture as the
		HD can achieve $91.4\%$. It can also be noticed that all the
		networks that have Relu samples failed to optimize properly, which
		is attributed to the finding discussed in \cite{Pennington2017} in which
		deep sigmoidal networks can outperform deep
		Relu networks if initialized properly. Moreover, all the networks with SGD optimizer
		struggle to achieve good results, which can be attributed to the
		small number of epochs used for the training. Finally, all the
		promising models have Softmax activation for the output layer. The
		top three models with respect to HD for each number of layers
		$l\in\{2,3,4\}$ are highlighted in Table
		\ref{tab:The-sampled-results} and further used for performance
		evaluation using the test set $X_{\text{\sc{tst}}}$.
		
		\begin{table}[t]
			\caption{Average Performance of the Theft Detector. \label{tab:Average-Detection-Performance}}
			
			\center
			\scalebox{1.0}{
				
				\begin{tabular}{ccccc}
					\toprule 
					& MD1  & MD2  & MD3  & \cite{jokar2016electricity}\tabularnewline
					\midrule
					\midrule 
					DR (\%)  & \textbf{93.4}  & 92.5  & 92.7  & 94\tabularnewline
					\midrule 
					FA (\%)  & 6.9  & \textbf{5.0}  & 5.3  & 11\tabularnewline
					\midrule 
					HD (\%)  & 86.4  & \textbf{87.4}  & 87.4  & 83\tabularnewline
					\bottomrule
				\end{tabular}
				
			}
			
		\end{table}

		Table \ref{tab:Average-Detection-Performance} gives a comparison
		between the top three models highlighted in Table
		\ref{tab:The-sampled-results} and the model proposed in
		\cite{jokar2016electricity}, which was tested by the same data used in this paper. The first three columns in Table \ref{tab:Average-Detection-Performance} give the average performance results when
		using the best hyper-parameters highlighted in Table
		\ref{tab:The-sampled-results}, such that the models with $l=2$, $l=3$ , and $l=4$ are represented by MD1, MD2, and MD3, respectively. The last column gives the results from
		\cite{jokar2016electricity}. While the detector in
		\cite{jokar2016electricity} is trained on the same dataset used in
		this paper, it is a customer-specific electricity theft detector
		based on a shallow machine learning architecture (SVM). As indicated in
		Table \ref{tab:The-sampled-results}, the best HD is achieved when
		MD2 or MD3 hyper-parameters are applied, which outperforms the proposed detector in
		\cite{jokar2016electricity} by a $4\%$ increase.
		Moreover, the FA is improved by almost $55\%$. The results indicate
		that an improvement in the electricity theft detection performance
		can be achieved using RNN architecture. In addition, our proposed
		detector is a general model that does not rely on a specific
		customer data. Hence, it is more robust against contamination
		attacks compared with the proposed detector in \cite{jokar2016electricity}.
		
		\vspace{-2bp}
		\section{Conclusion \label{sec:Conclusion}}
		\vspace{-1bp}
		A general RNN-based electricity theft detector is proposed. The
		proposed detector exploits the time series nature of the
		customers\textquoteright{} energy consumption to implement GRU deep
		hidden layers that can learn complex patterns of
		customers\textquoteright{} consumption yielding better detection
		performance. Random search is employed to tune the detector
		hyper-parameters and further improve the performance. Load profiles
		comprising $107,200$ days for different customers over two years are
		the basis for the experiments. The proposed RNN-based detector
		achieves a detection rate up to $93\%$ and a false acceptance rate
		as low as $5\%$, which presents a $55\%$ reduction in false
		acceptance, and $4\%$ improvement in the highest difference rate
		compared with a benchmark detector in the literature. Moreover, the
		proposed detector is more robust against contamination attacks as it
		does not rely on a specific customer data.
		
		\vspace{-5bp}
		\section{Acknowledgment}
		\vspace{-1bp}
		This publication was made possible by NPRP grant \# NPRP9-055-2-022
		from the Qatar National Research Fund (a member of Qatar
		Foundation). The statements made herein are solely the
		responsibility of the authors.
		\vspace{-7bp}
		\bibliographystyle{IEEEtran}
		\bibliography{biliography}

\begin{thebibliography}{10}
\providecommand{\url}[1]{#1}
\csname url@samestyle\endcsname
\providecommand{\newblock}{\relax}
\providecommand{\bibinfo}[2]{#2}
\providecommand{\BIBentrySTDinterwordspacing}{\spaceskip=0pt\relax}
\providecommand{\BIBentryALTinterwordstretchfactor}{4}
\providecommand{\BIBentryALTinterwordspacing}{\spaceskip=\fontdimen2\font plus
\BIBentryALTinterwordstretchfactor\fontdimen3\font minus
  \fontdimen4\font\relax}
\providecommand{\BIBforeignlanguage}[2]{{%
\expandafter\ifx\csname l@#1\endcsname\relax
\typeout{** WARNING: IEEEtran.bst: No hyphenation pattern has been}%
\typeout{** loaded for the language `#1'. Using the pattern for}%
\typeout{** the default language instead.}%
\else
\language=\csname l@#1\endcsname
\fi
#2}}
\providecommand{\BIBdecl}{\relax}
\BIBdecl

\bibitem{jokar2016electricity}
P.~Jokar, N.~Arianpoo, and V.~C. Leung, ``Electricity theft detection in {AMI}
  using customers' consumption patterns,'' \emph{IEEE Transactions on Smart
  Grid}, vol.~7, no.~1, pp. 216--226, 2016.

\bibitem{singh2017entropy}
S.~K. Singh, R.~Bose, and A.~Joshi, ``Entropy-based electricity theft detection
  in {AMI} network,'' \emph{IET Cyber-Physical Systems: Theory \&
  Applications}, 2017.

\bibitem{antmann2009reducing}
P.~Antmann, ``Reducing technical and non-technical losses in the power
  sector,'' \emph{World Bank, Washington, DC}, 2009.

\bibitem{dpreviewstaff}
\BIBentryALTinterwordspacing
P.~Pickering, ``{E-Meters Offer Multiple Ways to Combat Electricity Theft and
  Tampering},'' Last accesed: Nov 2017. [Online]. Available:
  \url{http://www.electronicdesign.com/meters/e-meters-offer-multiple-ways-combat-electricity-theft-and-tampering}
\BIBentrySTDinterwordspacing

\bibitem{nizar2008power}
A.~Nizar, Z.~Dong, and Y.~Wang, ``Power utility nontechnical loss analysis with
  extreme learning machine method,'' \emph{IEEE Transactions on Power Systems},
  vol.~23, no.~3, pp. 946--955, 2008.

\bibitem{nagi2011improving}
J.~Nagi, K.~S. Yap, S.~K. Tiong, S.~K. Ahmed, and F.~Nagi, ``Improving
  {SVM}-based nontechnical loss detection in power utility using the fuzzy
  inference system,'' \emph{IEEE Transactions on power delivery}, vol.~26,
  no.~2, pp. 1284--1285, 2011.

\bibitem{ramos2011new}
C.~Ramos, A.~de~Sousa, J.~Papa, and X.~Falcao, ``A new approach for
  nontechnical losses detection based on optimum-path forest,'' \emph{IEEE
  Transactions on Power Systems}, vol.~26, no.~1, pp. 181--189, 2011.

\bibitem{angelos2011detection}
E.~Angelos, O.~Saavedra, C.~Cort{\'e}s, and A.~de~Souza, ``Detection and
  identification of abnormalities in customer consumptions in power
  distribution systems,'' \emph{IEEE Transactions on Power Delivery}, vol.~26,
  no.~4, pp. 2436--2442, 2011.

\bibitem{lin2014non}
C.-H. Lin, S.-J. Chen, C.-L. Kuo, and J.-L. Chen, ``Non-cooperative game model
  applied to an advanced metering infrastructure for non-technical loss
  screening in micro-distribution systems,'' \emph{IEEE Transactions on Smart
  Grid}, vol.~5, no.~5, pp. 2468--2469, 2014.

\bibitem{amin2015game}
S.~Amin, G.~A. Schwartz, A.~A. Cardenas, and S.~S. Sastry, ``Game-theoretic
  models of electricity theft detection in smart utility networks: Providing
  new capabilities with advanced metering infrastructure,'' \emph{IEEE Control
  Systems}, vol.~35, no.~1, pp. 66--81, 2015.

\bibitem{zhou2015dynamic}
Y.~Zhou, X.~Chen, A.~Y. Zomaya, L.~Wang, and S.~Hu, ``A dynamic programming
  algorithm for leveraging probabilistic detection of energy theft in smart
  home,'' \emph{IEEE Transactions on Emerging Topics in Computing}, vol.~3,
  no.~4, pp. 502--513, 2015.

\bibitem{liu2015cyberthreat}
Y.~Liu and S.~Hu, ``Cyberthreat analysis and detection for energy theft in
  social networking of smart homes,'' \emph{IEEE Transactions on Computational
  Social Systems}, vol.~2, no.~4, pp. 148--158, 2015.

\bibitem{jindal2016decision}
A.~Jindal, A.~Dua, K.~Kaur, M.~Singh, N.~Kumar, and S.~Mishra, ``Decision tree
  and {SVM}-based data analytics for theft detection in smart grid,''
  \emph{IEEE Transactions on Industrial Informatics}, vol.~12, no.~3, pp.
  1005--1016, 2016.

\bibitem{bhat2016identifying}
R.~R. Bhat, R.~D. Trevizan, R.~Sengupta, X.~Li, and A.~Bretas, ``Identifying
  nontechnical power loss via spatial and temporal deep learning,'' in
  \emph{Proc. of 15th IEEE International Conference on Machine Learning and
  Applications (ICMLA)}, 2016, pp. 272--279.

\bibitem{zhan2016non}
T.-S. Zhan, S.-J. Chen, C.-C. Kao, C.-L. Kuo, J.-L. Chen, and C.-H. Lin,
  ``Non-technical loss and power blackout detection under advanced metering
  infrastructure using a cooperative game based inference mechanism,''
  \emph{IET Generation, Transmission \& Distribution}, vol.~10, no.~4, pp.
  873--882, 2016.

\bibitem{tariq2016electricity}
M.~Tariq and H.~V. Poor, ``Electricity theft detection and localization in
  grid-tied microgrids,'' \emph{IEEE Transactions on Smart Grid}, 2016.

\bibitem{xiao2017electricity}
F.~Xiao and Q.~Ai, ``Electricity theft detection in smart grid using random
  matrix theory,'' \emph{IET Generation, Transmission \& Distribution}, 2017.

\bibitem{bergstra2012random}
J.~Bergstra and Y.~Bengio, ``Random search for hyper-parameter optimization,''
  \emph{Journal of Machine Learning Research}, vol.~13, no. Feb, pp. 281--305,
  2012.

\bibitem{nagi2010nontechnical}
J.~Nagi, K.~S. Yap, S.~K. Tiong, S.~K. Ahmed, and M.~Mohamad, ``Nontechnical
  loss detection for metered customers in power utility using support vector
  machines,'' \emph{IEEE transactions on Power Delivery}, vol.~25, no.~2, pp.
  1162--1171, 2010.

\bibitem{he2008adasyn}
H.~He, Y.~Bai, E.~A. Garcia, and S.~Li, ``Adasyn: Adaptive synthetic sampling
  approach for imbalanced learning,'' in \emph{Proc. of IEEE International
  Joint Conference on Computational Intelligence}, 2008, pp. 1322--1328.

\bibitem{goodfellow2016deep}
I.~Goodfellow, Y.~Bengio, and A.~Courville, \emph{Deep learning}.\hskip 1em
  plus 0.5em minus 0.4em\relax MIT press, 2016.

\bibitem{dataset}
\BIBentryALTinterwordspacing
``{Irish Social Science Data Archive},'' Last accesed: Nov 2017. [Online].
  Available:
  \url{http://www.ucd.ie/issda/data/commissionforenergyregulationcer/}
\BIBentrySTDinterwordspacing

\bibitem{Pennington2017}
J.~Pennington, S.~Schoenholz, and S.~Ganguli, ``Resurrecting the sigmoid in
  deep learning through dynamical isometry: theory and practice,'' in
  \emph{Advances in neural information processing systems}, 2017, pp.
  4788--4798.

\end{thebibliography}

	\end{document}